\documentclass{article}
\usepackage{spconf,amsmath,graphicx}

\usepackage{enumitem}
\setlist{nosep, leftmargin=14pt}

\usepackage{multirow}
\usepackage{multicol}
\usepackage{algorithm2e}
\usepackage{xurl}
\usepackage{makecell}
\usepackage{algcompatible}
\usepackage{subfig}

\usepackage{xcolor}%

\title{Towards Out-of-Distribution Detection for breast cancer classification in Point-of-Care Ultrasound Imaging}
\name{\begin{tabular}{c} Jennie Karlsson$^{\star}$ \qquad Marisa Wodrich$^{\star}$ \qquad Niels Christian Overgaard$^{\star}$ \qquad Freja Sahlin$^{\star}$ \\
Kristina Lång$^{\dagger \ddagger}$ \qquad Anders Heyden$^{\star}$ \qquad Ida Arvidsson$^{\star}$\end{tabular}}

\address{$^{\star}$ Centre for Mathematical Sciences, Lund University, Sweden \\ $^{\dagger}$ Department of Diagnostic Radiology, Translational Medicine, Lund University, Sweden \\ $^{\ddagger}$ Unilabs Mammography Unit, Skåne University Hospital, Sweden}

\begin{document}
%
\maketitle

\begin{abstract}
Deep learning has shown to have great potential in medical applications. In critical domains as such, it is of high interest to have trustworthy algorithms which are able to tell when reliable assessments cannot be guaranteed. Detecting out-of-distribution (OOD) samples is a crucial step towards building a safe classifier. Following a previous study, showing that it is possible to classify breast cancer in point-of-care ultrasound images, this study investigates OOD detection using three different methods: softmax, energy score and deep ensembles. All methods are tested on three different OOD data sets. The results show that the energy score method outperforms the softmax method, performing well on two of the data sets. The ensemble method is the most robust, performing the best at detecting OOD samples for all three OOD data sets.
\end{abstract}

\begin{keywords}
Out-of-distribution detection, breast cancer, point-of-care ultrasound
\end{keywords}

\section{Introduction}
\label{sec:intro}
\begin{figure}[t!]
    \centering
\subfloat{\includegraphics[width=0.14\textwidth]{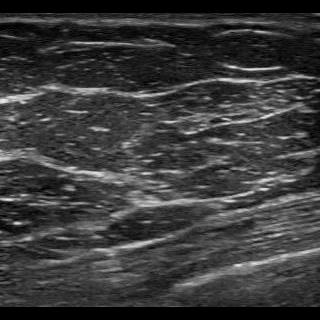}}\hspace{0.1mm}    \subfloat{\includegraphics[width=0.14\textwidth]{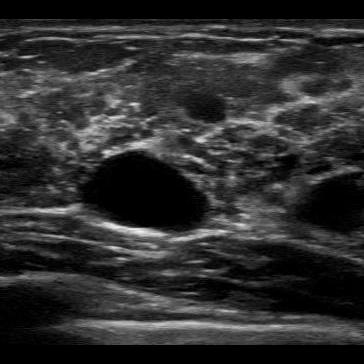}}\hspace{0.1mm}
\subfloat{\includegraphics[width=0.14\textwidth]{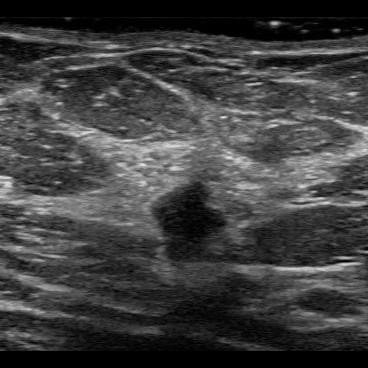}}\\
\vspace*{-3mm}
    \caption{POCUS images capturing normal tissue, benign and malignant lesions (from the left to right).}
    \label{fig:POCUS_ex}
\end{figure}

Deep learning has achieved promising results assessing different types of medical images~\cite{zhou2021review}. However, for safe deployment and trustworthiness, the algorithms have to be able to tell when they cannot make proper assessments. This includes detecting out-of-distribution (OOD) data, which comprise data samples that the algorithm has not learned how to interpret. A lot of progress has been made in this area in recent years with numerous methods having been suggested for OOD detection~\cite{survey} and uncertainty quantification~\cite{nemani2023uncertainty}. In this study we explore different methods for OOD detection, including uncertainty-based ones, with the aim of being used in a tool for breast cancer classification in point-of-care ultrasound (POCUS) images. 
Breast cancer is the most common type of cancer amongst women worldwide \cite{sung2021global}. Detecting breast cancer in an early stage improves patient outcome both in terms of mortality and morbidity, but access to diagnostics is lacking in many low- and middle-income countries~\cite{francies2020breast, GBCI}. A possible solution could be a deep learning-based algorithm analyzing images captured with a POCUS device and using a smart-phone for visualization. Examples of POCUS images capturing breast tissue are shown in Fig.~\ref{fig:POCUS_ex}. In a previous study it has been shown that a convolutional neural network (CNN) can classify breast cancer in POCUS images with good performance~\cite{spie2023}. Here we extend this work by exploring and evaluating three different OOD detection methods: softmax, energy score and deep ensembles. This should enable detecting unsuitable images during inference, which is crucial to make reliable assessments in a real world setting. 

\section{Theory}
\subsection{Softmax}
The softmax probabilities from the last layer of a neural network have previously been used for OOD detection~\cite{hendrycks2016baseline}. The idea is that samples with low softmax scores for the predicted class are identified as OOD. However, an issue with softmax is that the scores for the classes need to add up to one, even when none of the classes fits the sample. In this study, softmax has been used as a baseline.

\subsection{Energy score}
Energy score is a post-hoc OOD detection method using the logits, i.e. the unscaled outputs from the network before the softmax activation. By looking at the logits, the issue of the softmax score requirement to add up to one is avoided. The method was proposed in~\cite{liu2020energy} and does not require any retraining of the neural network. For input $x$ and network $f(x)$, the energy score can be expressed as
\begin{equation}
    E(x;f) = -T \cdot \log\sum_{i}^{K}e^{f_i(x)/T},
    \label{eq:energy}
\end{equation}
where the logit for class label $i$ and input $x$ is denoted by $f_i(x)$, $K$ denotes the total number of class labels and $T$ is a temperature parameter. 
A related work to energy score is Multi-Level Out-of-Distribution Detection (MOOD)~\cite{lin2021mood}. This method also uses energy score but instead of just analysing the score in the end of the network, it analyse multiple exits throughout the network. The idea is that obvious OOD samples should be detected earlier in the network and more complex samples should be detected further in into the network.

\subsection{Deep ensembles}
Deep neural network ensembling~\cite{hansen1990neural} is a method to improve the generalizability and reliability of a network by independently training multiple networks on the same problem. The combined output makes it possible to calculate uncertainties by looking at the differences between the separate predictions. These uncertainties can be used for OOD detection. 

Measuring uncertainty in an ensemble can be achieved in various ways, for instance by calculating the standard deviation of the predictions. For OOD data samples, the uncertainty should be high as opposed to in-distribution (ID) data, where the epistemic uncertainty should be small~\cite{lakshminarayanan2017simple}.

\section{Data}
\subsection{In-distribution data}
The ID data consists of POCUS images capturing breast tissue collected with a GE Vscan air probe~\cite{GE} at Skåne University Hospital, Malmö. This data set contains images of normal tissue as well as benign and malignant lesions, see Fig.~\ref{fig:POCUS_ex}. The data was split into training and test set. In addition to the POCUS training set, a conventional ultrasound (US) data set of breast tissue was also used for training. These images were collected with Logiq E9 and Logiq E10 ultrasound machines at Skåne university hospital, Malmö. The sizes of the ID data sets are shown in Table~\ref{tab:POCUS}.

\begin{table}[t!]
    \footnotesize
    \centering
    \caption{The sizes of the ID data sets.}
    \vspace*{2mm}
    \begin{tabular}{c c c c}
    \hline
     & Train (POCUS) & Train (US) &Test (POCUS)\\\hline
    Normal & 304 & 168 & 284\\
    Benign & 140 & 101 & 131\\
    Malignant & 125 & 398 & 116\\\hline
    Total & 569 & 667 & 531\\\hline
    \end{tabular}
    \label{tab:POCUS}
\end{table}

\subsection{Out-of-distribution data}
\begin{figure}
    \centering
\subfloat{\includegraphics[width=0.11\textwidth]{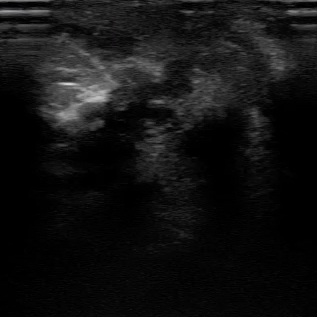}}\hspace{0.1mm} 
\subfloat{\includegraphics[width=0.11\textwidth]{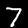}}\hspace{0.1mm}
\subfloat{\includegraphics[width=0.11\textwidth]{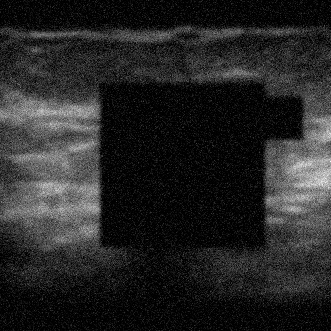}}\hspace{0.1mm} 
\subfloat{\includegraphics[width=0.11\textwidth]{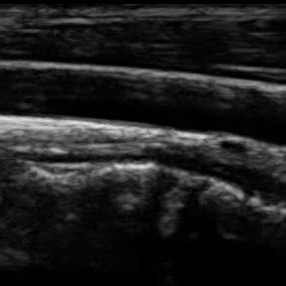}}\\
\vspace*{-3mm}
\caption{Example of a POCUS image of poor quality, and images from the OOD data sets MNIST, CorruptPOCUS and CCA (left to right).}
\label{fig:OOD_examples}
\end{figure}

To evaluate the performance in terms of OOD detection, three different OOD test data sets were used: MNIST (test set), CorruptPOCUS and CCA. The MNIST test set consists of 10~000 images of handwritten digits~\cite{deng2012mnist}. The CorruptPOCUS and CCA data sets were chosen to resemble realistic OOD data, such as ultrasound images containing artefacts. The CorruptPOCUS data set resembles POCUS images of poor quality and was generated by distorting the POCUS test set by adding dark areas, blur and noise. The CCA data set contains 84 ultrasound images capturing the common carotid artery~\cite{cca}, i.e., non-breast US images.
An example of a real POCUS image of poor quality as well as examples of the three OOD data sets can be seen in Fig.~\ref{fig:OOD_examples}.

\section{Methods}
\label{sec:cnn}
\subsection{Classification network}
The classification network was implemented as a CNN consisting of five convolutional layers followed by two fully connected layers. The CNN architecture, training settings, training data and augmentation is described in~\cite{spie2023}.

\subsection{Softmax}
For the softmax OOD detection method, the classification network above was used. The softmax scores were obtain from the output of this CNN.

\subsection{Energy score}
\begin{figure}[tbp]
\centerline{\includegraphics[width=0.5\textwidth]{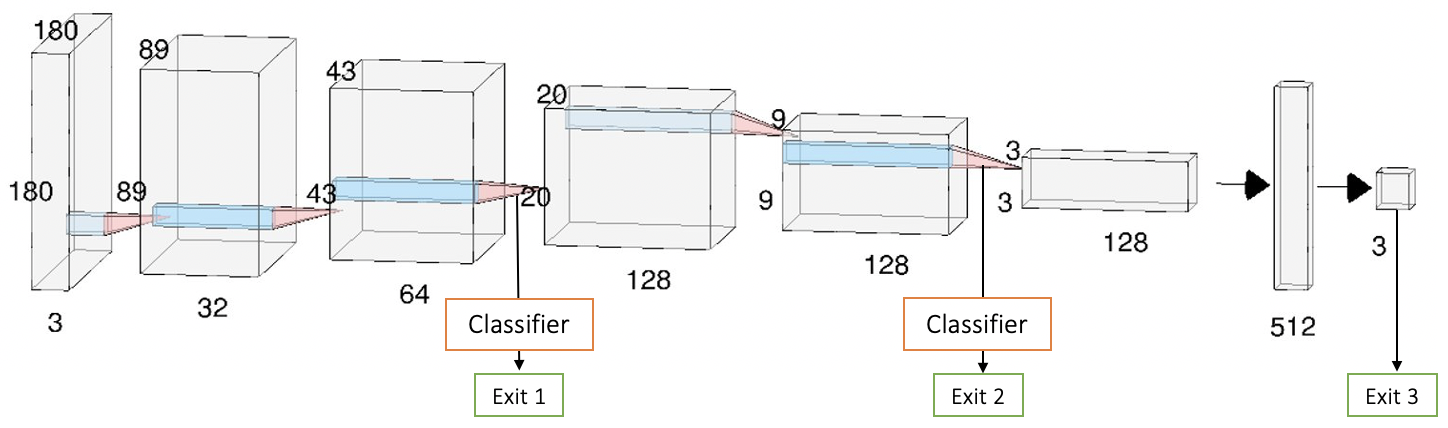}}
\vspace*{-2mm}
\caption{Architecture of CNN with additional classifiers and exits. The scheme was created in NN-SVG~\cite{LeNail2019}.}
\label{fig:cnn_exits}
\end{figure}

The energy score was implemented according to Eq.~(\ref{eq:energy}). After trying out different values for the temperature parameter $T$ based on the ID data, it was set to 0.001. The energy score was obtained from the classification network, which was modified to have three exits from different parts of the network. The modified architecture of the CNN is displayed in Fig.~\ref{fig:cnn_exits}. The classifiers added to the first two exits consisted of one convolutional layer with 128 kernels and kernel size 3x3, a ReLU activation, a 2x2 max pooling and a fully connected layer of size three without any activation. The CNN was trained with three categorical cross entropy losses, using the outputs from the first two exits and the final softmax output. The three losses were weighted 0.5, 0.5, 1, making the last exit influence the training the most.  For OOD detection, thresholds were found for each exit and if a sample had higher energy than all three thresholds it was labeled as ID, otherwise OOD.

\subsection{Deep ensemble}
For the ensemble, 20 independent models with the above described classification network architecture were trained. In order to diversify the models, we left out random 0-15\% of the training data for each of them and randomly chose some of the training hyperparameters. The learning rate was between 0.0001 and 0.001, optimizer  was either Adam or RMSprop, number of epochs between 25 and 85 and the batch size was either 8, 16, 32, 64 or 128.

As the measure for uncertainty we chose the standard deviation. Two versions were implemented, one using the sum of standard deviations per image for all three classes, and the second one weighting them by the mean of the classes before summation. For classification, majority vote within the ensemble was used.

\subsection{Metrics}
Three types of metrics were used to evaluate the different methods: receiver operating characteristic  (ROC), area under the ROC curve (AUC) and false positive rate (FPR). For the OOD detection both ROC and AUC were evaluated, using the POCUS test set as ID and the OOD sets as OOD. The FPR was evaluated at the OOD classification threshold where 95\% of the ID data was classified as ID, referred to as FPR95. For evaluating the classification of the POCUS test set into normal and benign (non-cancerous) versus malignant (cancerous), AUC was used. For ensembles, the AUC was calculated with the average output for the malignant class. Additionally, the AUC was obtained when letting 5\% of the POCUS test set be detected as OOD, referred to as FNR5. 

\section{Results}
\begin{figure*}[tb]
    \centering
    \subfloat{\includegraphics[width=0.24\textwidth]{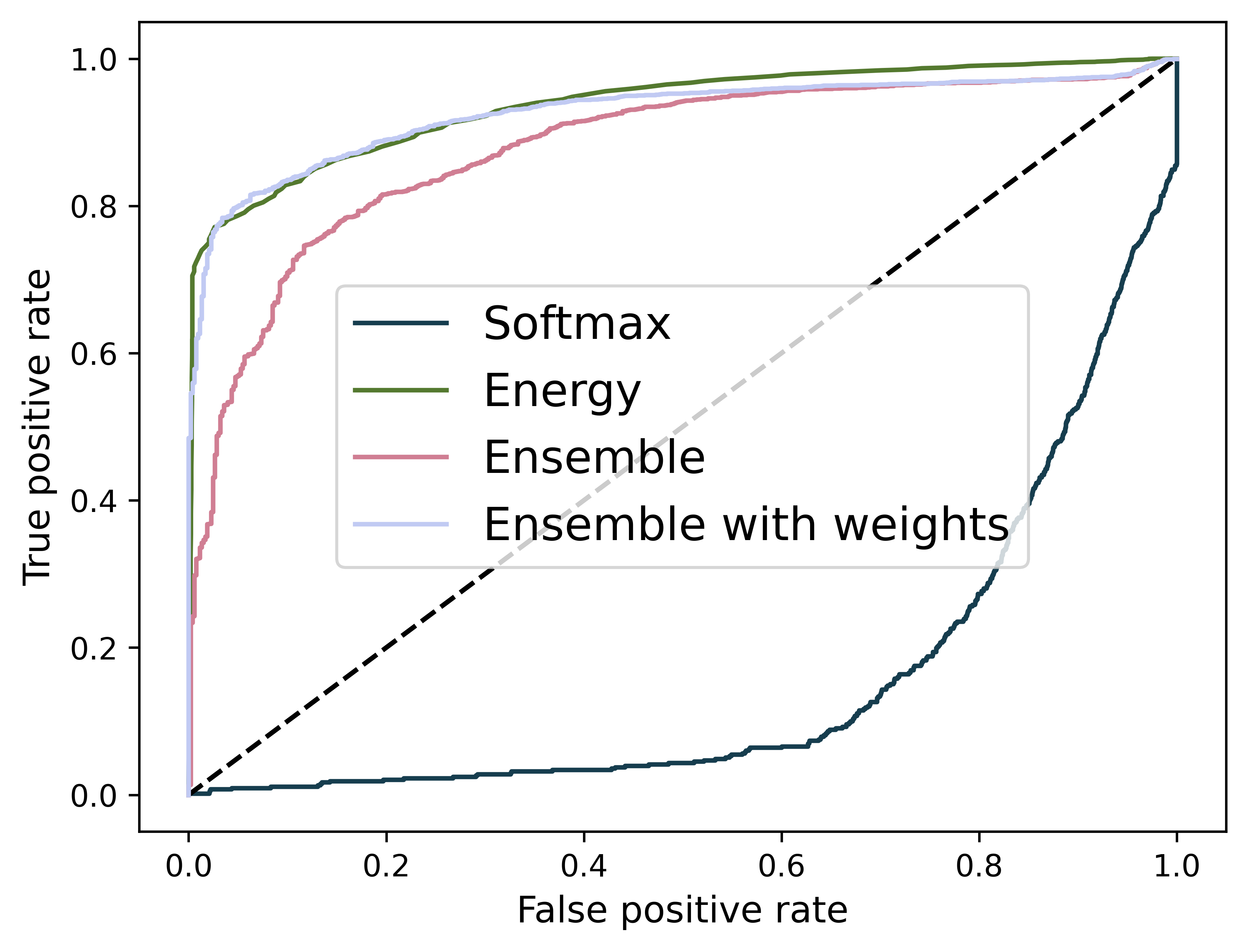}}\hspace{0.1mm}    \subfloat{\includegraphics[width=0.24\textwidth]{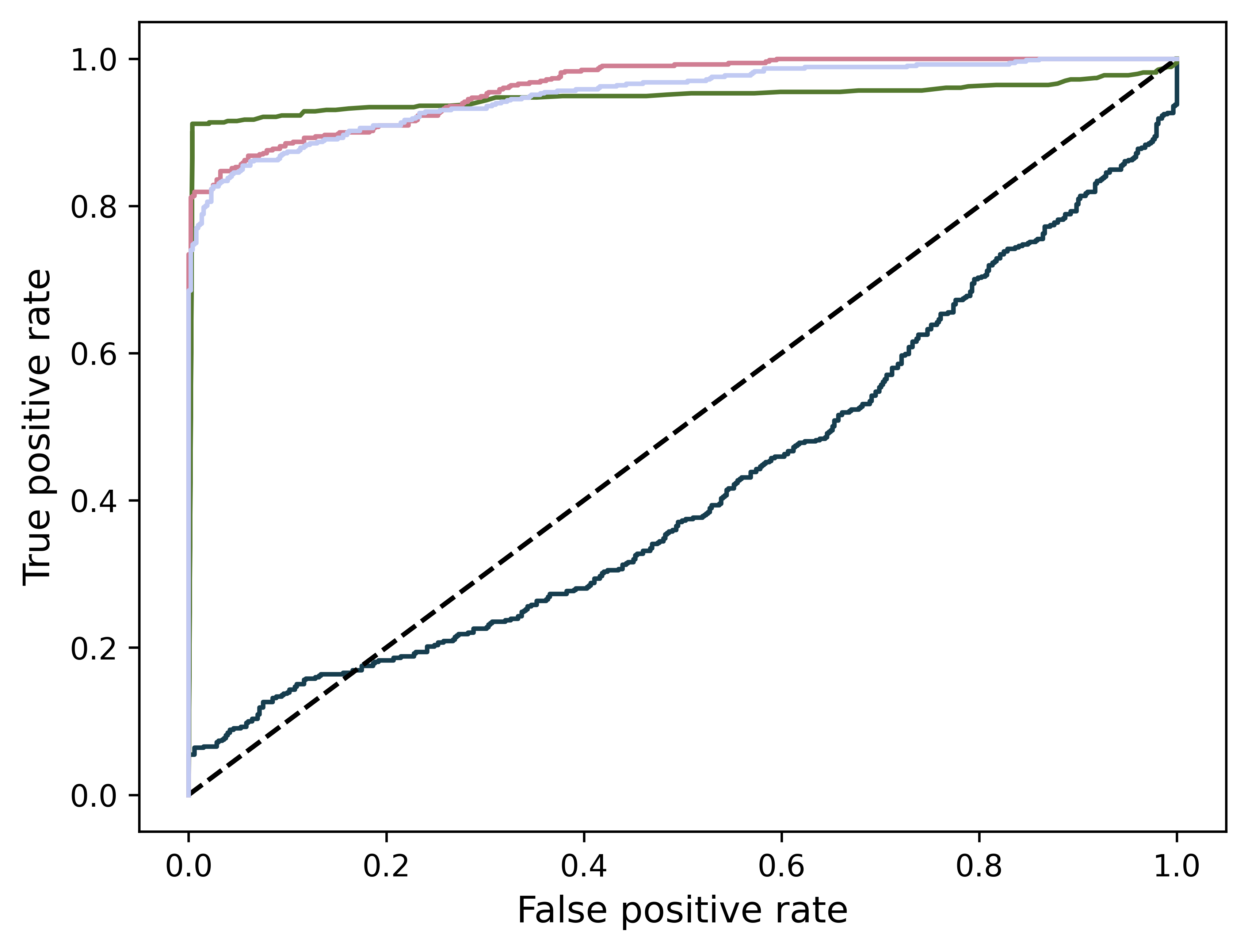}}\hspace{0.1mm}
    \subfloat{\includegraphics[width=0.24\textwidth]{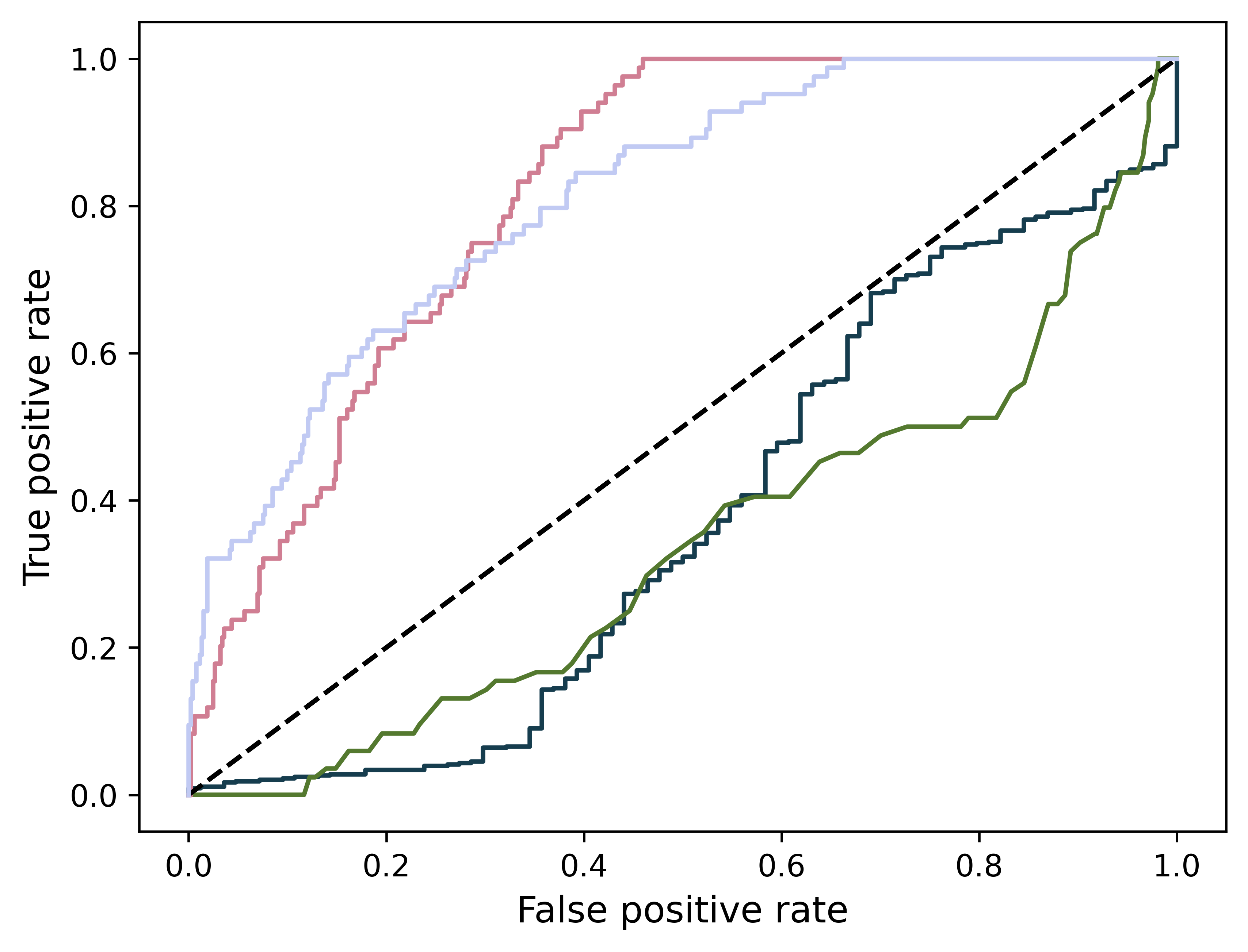}}\\
    \vspace*{-3mm}
    \caption{ROC curves for the OOD detection methods evaluated on MNIST (left), CorruptPOCUS (middle) and CCA (right).}
    \label{fig:roc}
\end{figure*}

\begin{figure}[tb]
    \centering
\subfloat{\includegraphics[width=0.45\textwidth]{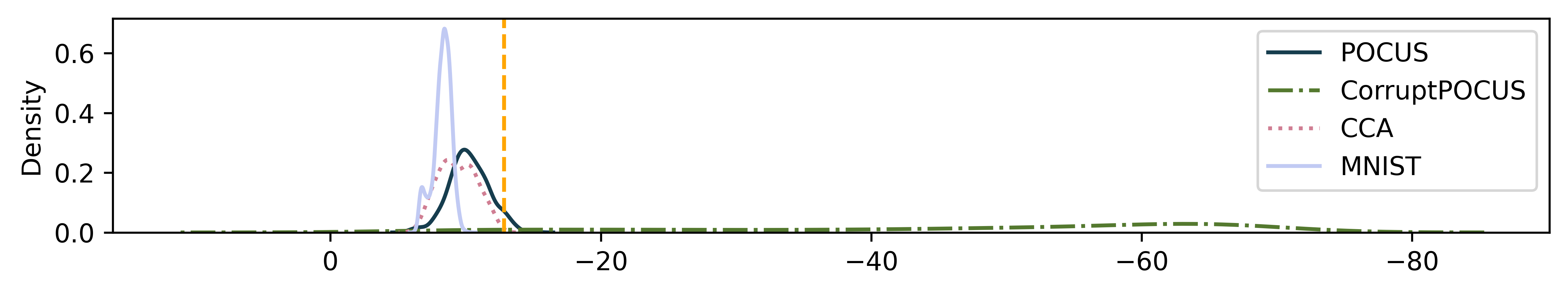}}\\  \subfloat{\includegraphics[width=0.45\textwidth]{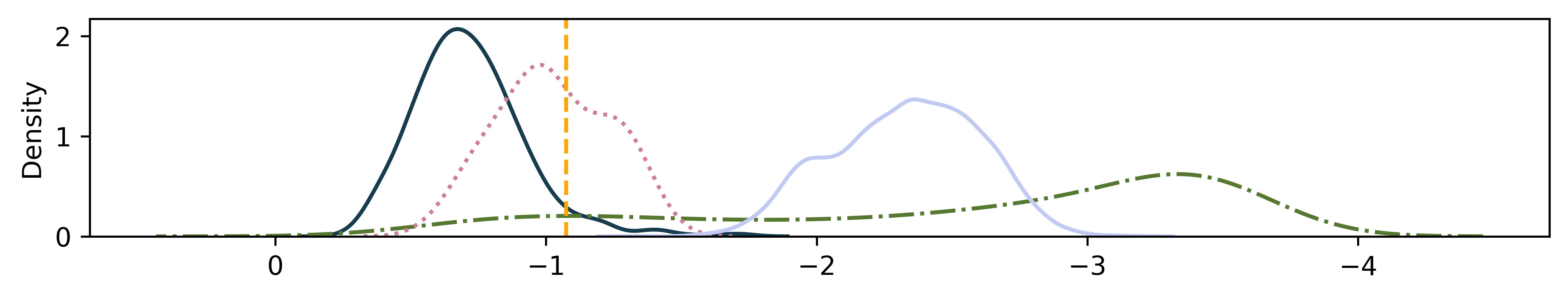}}\\
\subfloat{\includegraphics[width=0.45\textwidth]{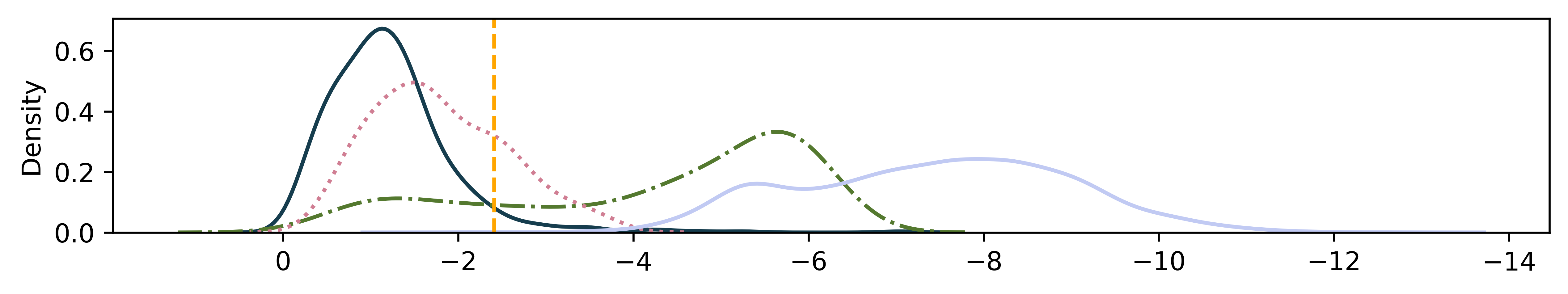}}\\
\vspace*{-3mm}
    \caption{Distribution of energy scores for the OOD data sets. Energies from exit 1 (top), exit 2 (middle) and exit 3 (bottom). The vertical line marks the threshold where 95\% of the POCUS test set images are classified as ID. }
    \label{fig:energy_dist}
\end{figure}

\begin{figure}[tb]
    \centering
    \includegraphics[width=0.45\textwidth]{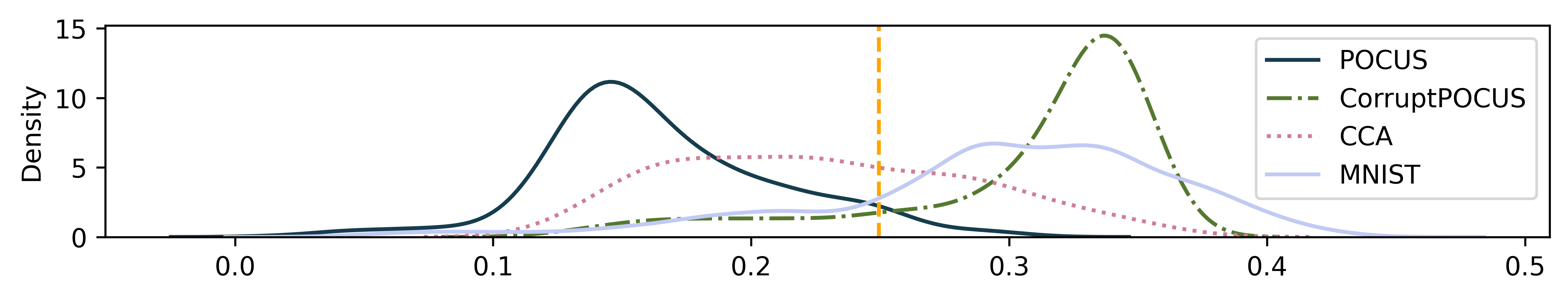}
    \includegraphics[width=0.45\textwidth]{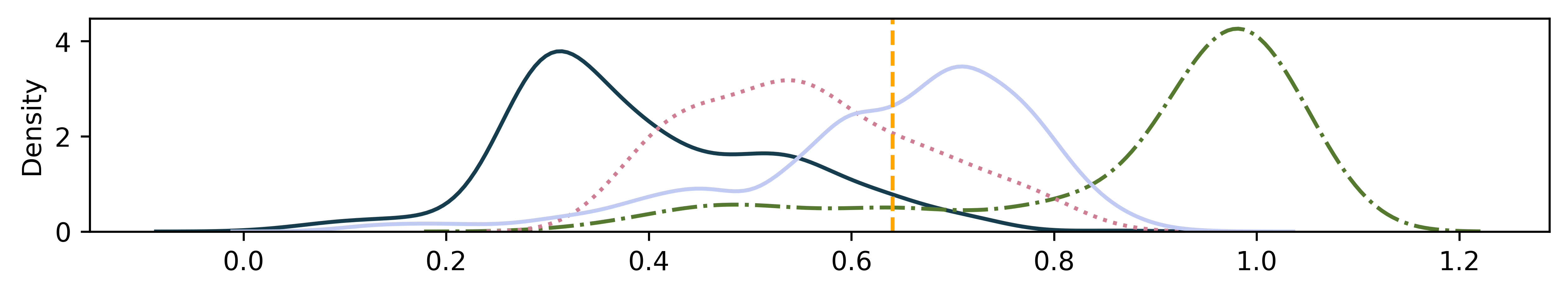}
    \vspace*{-3mm}
    \caption{Distribution of weighted (top) and unweighted (bottom) epistemic uncertainty of ensemble for the different OOD sets. The vertical line marks the threshold where 95\% of the POCUS test set images are classified as ID.}
    \label{fig:ens_dist}
\end{figure}

The AUC and FPR95 was evaluated for all OOD detection methods for each OOD data set, the results are displayed in Table~\ref{tab:res_OOD}. The OOD detection methods were also evaluated by plotting the ROC curves for each OOD data set, Fig.~\ref{fig:roc}. The individual results for OOD detection using energy scores from the different exits can be seen in Table~\ref{tab:exits_results}. To visualise the energies from the energy score method, the distribution for each OOD data set and exit is shown in Fig.~\ref{fig:energy_dist}. Corresponding plots for the epistemic uncertainties from the ensemble method can be seen in Fig.~\ref{fig:ens_dist}. Finally the AUC and AUC at FNR5 for classification of cancer in the POCUS test set was calculated for each OOD method, displayed in Table~\ref{tab:res_cancer}. 

\begin{table}[tb]
    \footnotesize
    \centering
    \caption{AUC and FPR for the different OOD detection methods evaluated on the OOD data sets. Here $\downarrow$ implies smaller values are superior and $\uparrow$ implies larger values are superior.}
    \vspace*{2mm}
    \begin{tabular}{ccccc}
    \hline
    Method & OOD data & AUC (\%) $\uparrow$ & FPR95 (\%) $\downarrow$\\\hline
    \multirow{ 3}{*}{Softmax} & MNIST & 15.6 & 100.0\\
    & CorruptPOCUS & 42.1 & 100.0\\
    & CCA & 37.4 & 100.0\\\hline
    \multirow{ 3}{*}{Energy} & MNIST & 93.5 & 21.4\\
    & CorruptPOCUS & 94.6 & 8.5\\
    & CCA & 33.8 & 100.0\\\hline
    \multirow{ 3}{*}{Ensemble} & MNIST & 87.8  & 43.1\\
    & CorruptPOCUS & 96.3 & 14.7\\
    & CCA & 81.8 & 76.2\\\hline
     \multirow{ 3}{*}{\makecell{Ensemble \\ with weights}} & MNIST & 92.6  & 20.1\\
    & CorruptPOCUS & 95.2 & 15.4\\
    & CCA & 81.2 & 65.5\\\hline
    \end{tabular}
    \label{tab:res_OOD}
\end{table}

\begin{table}[tb]
    \centering
    \footnotesize
    \caption{AUC and FPR for each exit and OOD data set.}
    \vspace*{2mm}
    \begin{tabular}{c |c c c |c c c}
    \hline
     OOD data & \multicolumn{3}{c}{AUC (\%) $\uparrow$} & \multicolumn{3}{|c}{FPR95 (\%) $\downarrow$}\\\hline
      & \textbf{exit 1} & \textbf{exit 2} & \textbf{exit 3} & \textbf{exit 1} & \textbf{exit 2} & \textbf{exit 3}\\
       MNIST & 9.9 & 100.0 & 100.0 & 100.0 & 0.0 & 0.0\\
      CorruptPOCUS & 94.9 & 95.8 & 90.8 & 8.5 & 11.1 & 22.0\\
      CCA & 34.0 & 86.1 & 73.0 & 100.0 & 60.1 & 77.4\\\hline
    \end{tabular}
    \label{tab:exits_results}
\end{table}

\begin{table}[tb]
    \centering
    \footnotesize
    \caption{The different methods' results for classification of cancer versus non-cancer.}
    \vspace*{2mm}
    \begin{tabular}{ccccc}
    \hline
        Method & AUC (\%) $\uparrow$ & AUC at FNR5 (\%) $\uparrow$\\\hline 
        Softmax & 95.3 & 96.1\\ 
        Energy & 95.0 & 94.7\\ 
        Ensemble & 96.6 & 96.5\\ 
        Ensemble with weights & 96.6  & 96.1\\\hline 
    \end{tabular}
    \label{tab:res_cancer}
\end{table}

\section{Discussion}
The softmax method achieved inferior performance compared to the other methods on all data sets, with low AUC and a 100 \% FPR95 for all OOD data sets, see Table~\ref{tab:res_OOD}. It has previously been shown that the softmax output can be overconfident for samples far away from the training data~\cite{softmax_overconfident}, which our results corroborate. The energy scores from different exits seems to be useful for different OOD data, as can be seen in Table~\ref{tab:exits_results}. For example, MNIST samples are not well detected as OOD in the first exit, but in the last two exits they are perfectly detected as OOD. On the contrary the CorruptPOCUS data set is better detected in the first two exits. The CCA data set is detected as OOD  poorly for all exits. Compared to the softmax and energy score methods, ensemble is the only method that perform well on all three OOD data sets. This can be seen in Table~\ref{tab:res_OOD} and in the ROC curves in Fig.~\ref{fig:roc}. 
The results for the ensemble with weights are similar or slightly better than the unweighted ones for all four data sets, which is in alignment with the results shown in Fig.~\ref{fig:ens_dist}. Using weights for the uncertainties makes it easier to separate the distributions of ID and OOD data.

The softmax and energy score methods have the advantages of not requiring any re-training of the network, hence are easy to implement. Ensembles come with the drawback of complexity, in our case having a 20 times higher training and inference time compared to the other methods. 

According to Table~\ref{tab:res_OOD} and Table~\ref{tab:exits_results}, the CCA data set is poorly detected as OOD. These images are ultrasound images just like the ID data, with the difference that they are capturing the common carotid artery instead of breast tissue. Their appearance might be too close to the ID data. Thus, the OOD detection methods used in this paper do not manage to identify images that cannot be classified reliably, such as images from other types of tissue. This is also supported by the results in Table~\ref{tab:res_cancer}, where removing OOD samples with the OOD detection methods does not have any larger impact on the AUC. This implies that the performance of the classifiers seems to be the same in all regions of data that lie within the same distribution as the training data, independent of the OOD detection method.

Even though the OOD methods might not work as well for data very close to the ID data, it shows promising results detecting data capturing corrupted ultrasound images. In a real world setting this has the potential of being useful, by having the OOD detector flag when an image is of poor quality. 

For future research more OOD detection methods should be investigated, including Bayesian neural networks, deterministic uncertainty quantification methods and post-hoc OOD detection methods.

\section{Conclusion}
In this work three different OOD detection methods have been compared and evaluated. The ensemble methods had more robust results, performing well on all OOD data sets. Energy score performed well on the MNIST and CorruptPOCUS data sets but poorly on the CCA data set. The relative complexity of the ensemble method requires more computational power compared to energy score. The study illustrates the problem of finding a balance between performance and computational complexity when it comes to OOD detection. Finding OOD samples is important if deep learning is to be used in a real world medical setting. The methods show promising result for detecting OOD data far from the ID data, but further research is needed in order to detect OOD data very similar to the ID data.

\section{Compliance with ethical standards}
This study was performed in line with the principles of the Declaration of Helsinki. Approval was granted by the Swedish Ethical Review Authority of Region Skåne (2019-04607).

\section{Acknowledgments}
This work was supported by strategic research area eSSENCE and Analytic Imaging Diagnostics Arena (AIDA).

\bibliographystyle{IEEEbib}
\bibliography{strings,refs}

\end{document}